\documentclass[conference]{IEEEtran}
\IEEEoverridecommandlockouts
\usepackage{cite}
\usepackage{amsmath,amssymb,amsfonts}
\usepackage{graphicx}
\usepackage{booktabs}
\usepackage{tabularx}
\usepackage{makecell}
\usepackage{textcomp}
\usepackage{xcolor}
\usepackage{stfloats}
\usepackage{balance}
\usepackage[most]{tcolorbox}
\usepackage{fancyhdr}
\usepackage{hyperref}
\usepackage{algorithm}
\usepackage{algpseudocode}
\def\BibTeX{{\rm B\kern-.05em{\sc i\kern-.025em b}\kern-.08em
    T\kern-.1667em\lower.7ex\hbox{E}\kern-.125emX}}

\begin{document}
\title{Chat3GPP: An Open-Source Retrieval-Augmented Generation Framework for 3GPP Documents\\
}
\author{
	\IEEEauthorblockN{
		Long Huang\IEEEauthorrefmark{1}, 
		Ming Zhao\IEEEauthorrefmark{2}, 
		Limin Xiao\IEEEauthorrefmark{2}, 
		Xiujun Zhang\IEEEauthorrefmark{2},
		Jungang Hu\IEEEauthorrefmark{2}}
	\IEEEauthorblockA{\IEEEauthorrefmark{1}Dept of Electronic Engineering, Tsinghua University, Beijing, China}
	\IEEEauthorblockA{\IEEEauthorrefmark{2}Beijing National Research Center for Information Science and Technology, Beijing, China}
	\IEEEauthorblockA{Email: huangl22@mails.tsinghua.edu.cn, \{zhaoming,xiaolm,zhangxiujun,jungangh\}@tsinghua.edu.cn}
} 


\maketitle

\begin{abstract}
The 3rd Generation Partnership Project (3GPP) documents is key standards in global telecommunications, while posing significant challenges for engineers and researchers in the telecommunications field due to the large volume and complexity of their contents as well as the frequent updates. Large language models (LLMs) have shown promise in natural language processing tasks, but their general-purpose nature limits their effectiveness in specific domains like telecommunications. To address this, we propose Chat3GPP\hyperlink{footnote1}{\textsuperscript{1}}, an open-source retrieval-augmented generation (RAG) framework tailored for 3GPP specifications. By combining chunking strategies, hybrid retrieval and efficient indexing methods, Chat3GPP can efficiently retrieve relevant information and generate accurate responses to user queries without requiring domain-specific fine-tuning, which is both flexible and scalable, offering significant potential for adapting to other technical standards beyond 3GPP. We evaluate Chat3GPP on two telecom-specific datasets and demonstrate its superior performance compared to existing methods, showcasing its potential for downstream tasks like protocol generation and code automation.
\end{abstract}
\hypertarget{footnote1}{}
\footnotetext[1]{Open-source implemention is available at \url{https://github.com/huangl22/Chat3GPP}}
\section{Introduction}
In global telecommunications networks, the 3rd Generation Partnership Project (3GPP) documents serve as a foundational standard, covering various aspects such as wireless communication, network architecture, core networks, access networks, and signaling and data transmission between devices. Since its establishment in 1998, the 3GPP standards have evolved through multiple generations of communication technologies, including 2G, 3G, 4G, 5G, and continue to play an important role in the development of 6G standards. The extensive scope and complexity of 3GPP documents, filled with specialized terminology and subject to frequent updates and revisions, present significant challenges. Even experienced engineers often need to spend substantial time reviewing and comprehending these documents to grasp key details. In this context, how to efficiently and accurately understand and apply these specification documents has become a critical challenge for technicians, engineers, and researchers in the telecommunications field.

Recently, large language models (LLMs) have made significant developments and advancements in the field of natural language processing (NLP). Models such as LLaMA\cite{b1} and Mistral\cite{b2} have demonstrated outstanding performance in tasks such as text generation, summarization and question-answering, owing to their large-scale pre-training, which enables them to generate and reason with text more naturally and accurately. However, while these general-purpose LLMs perform well across many domains, they face limitations when applied to specialized fields like telecommunications. In order to bridge the gap between general-purpose LLMs and the specialized needs of the telecommunications, researchers have made substantial efforts. Bariah et al.\cite{b3} adapted pre-trained generative models for identifying the 3GPP standard working groups. Maatouk et al.\cite{b4} curated TeleQnA, a multiple-choice question (MCQ) benchmark dataset to evaluate the knowledge of LLMs in telecommunications. Furthermore, Zou et al.\cite{b5} introduced to adapt general-purpose LLMs to telecom-specific tasks, while Maatouk et al.\cite{b6} focued on open-sourcing a series of LLMs for telecommunications applications, along with two datasets for training and evaluation. 

However, training telecom-specific LLMs presents several challenges, including the lack of comprehensive telecommunications datasets and the diverse presentation formats such as figures and tables that complicate the training process. Additionally, the frequent updates in telecom knowledge make it difficult for LLMs to stay up-to-date, with high training costs as a further barrier. Therefore, a method, retrieval augmented generation (RAG), has emerged as an effective solution due to its cost effectiveness, adaptability, and scalability. RAG enhances the capability of LLMs by integrating an information retrieval system that supplies relevant data during the generation process. Recently, Telco-RAG\cite{b7} provides generally applicable guidelines for overcoming common challenges in implementing an RAG pipeline in highly technical domains. 

In this Paper, We introduce Chat3GPP, an RAG pipeline for 3GPP documents, which not only reduces model maintenance costs but also enhances the system’s flexibility and scalability. By constructing a 3GPP document embedding database, it can swiftly retrieve relevant information in response to user queries and generate answers through a model. Importantly, Chat3GPP is not limited to 3GPP documents and its framework can be easily extended to other standard documents, offering significant potential for broader application. The key contributions of this paper are as follows:
\begin{itemize}
    \item We propose Chat3GPP, an open-source chatbot for 3GPP documents based on RAG framework without any domain-specific fine-tuning. This allows for easy extension to other technical standards, providing a solid foundation for downstream tasks such as protocol generation and code automation.
    \item We evaluate Chat3GPP on open-source telecom datasets and show its superior performance compared to the existing methods.
\end{itemize}

The remainder of this paper is organized as follows. In the next section, we elaborate on the related work. In Section III we describe the methodology behind Chat3GPP, detailing the data pre-processing, indexing, retrieval, and generation processes. In Section IV, we present the experimental settings and evaluation results. Finally, We conclude the paper and suggest potential avenues for future work.

\section{Related Work}
This section first reviews the research progress on the application of LLMs in the telecommunications domain, and then introduces the background of RAG and summarizes the advancements in telecom-specific RAG research.

\subsection{LLMs in Telecommunications}
The application of large language models (LLMs) in the telecommunications domain has been a topic of increasing interest due to their potential to revolutionize various aspects of the industry. Maatouk et al.\cite{b8} explored the potential impact of LLMs in the telecommunications industry and examined their application scenarios in the telecommunications field, while pointing out the key research directions that need to be addressed to fully exploit the potential of LLMs. Bariah et al.\cite{b3} demonstrated the potential of LLMs for identifying 3GPP standard working groups by fine-tuning models such as BERT, DistilBERT, RoBERTa, and GPT-2. Nabeel et al.\cite{b9} introduced a new framework to generate test scripts using a hybrid generative model, reducing lead time in telecom software development. Zou el al.\cite{b5} proposed a framework for adapting general-purpose LLMs to telecom-specific LLMs by continuous pre-training, instruction tuning, and alignment tuning seperatly on pre-training datasets, instruction datasets, and preference datasets, demonstrating the effectiveness of domain-specific training and tuning. Maatouk et al.\cite{b6} focused on developing a series of LLMs specifically tailored for the telecommunications domain, known as Tele-LLMs. They created the Tele-Data and Tele-Eval datasets and conducted extensive experiments with various training techniques to identify the optimal strategies for adapting LLMs to the telecommunications field.

\subsection{RAG}
RAG is a technique that combines information retrieval with language generation models to enhance the performance of LLMs in knowledge-intensive tasks. RAG was introduced to addresses the limitations of LLMs initially\cite{b10}, such as knowledge cutoff and hallucination, by integrating external knowledge into the model. The core idea is to retrieve relevant information from a knowledge base and use it to augment the input to the language model, thereby improving the accuracy and relevance of the generated text.

The RAG system consists of two main components: the retriever and the generator. The retriever module uses a pre-trained text embedding model to convert queries and documents into vector representations, which are then searched in a embedding database to find the most relevant documents. The generator module combines the retrieved documents with the original query to form a richer context, which is used to generate the final response. This architecture has been applied in various domains, including question answering, text summarization, and content generation, where RAG has shown significant improvements in performance.

Recent research has focused on optimizing and extending the RAG framework. For instance, RAFT\cite{b11} adapts LLMs to specific domains by training the model to ignore irrelevant retrieved documents and focus on relevant information. Corrective RAG\cite{b12} enhances the robustness and precision of LLMs by evaluating the quality and relevance of retrieved documents and using a confidence-based adaptive retrieval mechanism. Additionally, RA-ISF\cite{b13} uses an iterative self-feedback process to enhance the problem-solving efficiency of LLMs, resulting in improved performance in factual reasoning and reduced hallucination.

The application of RAG in the telecommunications domain presents unique challenges due to the complex nature of telecom standard documents and the rapid evolution of the field. Telco-RAG\cite{b7}, an open-source RAG framework, has been developed to handle the specific needs of telecommunications standards, particularly 3GPP documents. Telco-RAG addresses the critical challenges of implementing a RAG pipeline on highly technical content, paving the way for applying LLMs in telecommunications and offering guidelines for RAG implementation in other technical domains.

\begin{figure*}[htbp]
\centerline{\includegraphics[width=\linewidth]{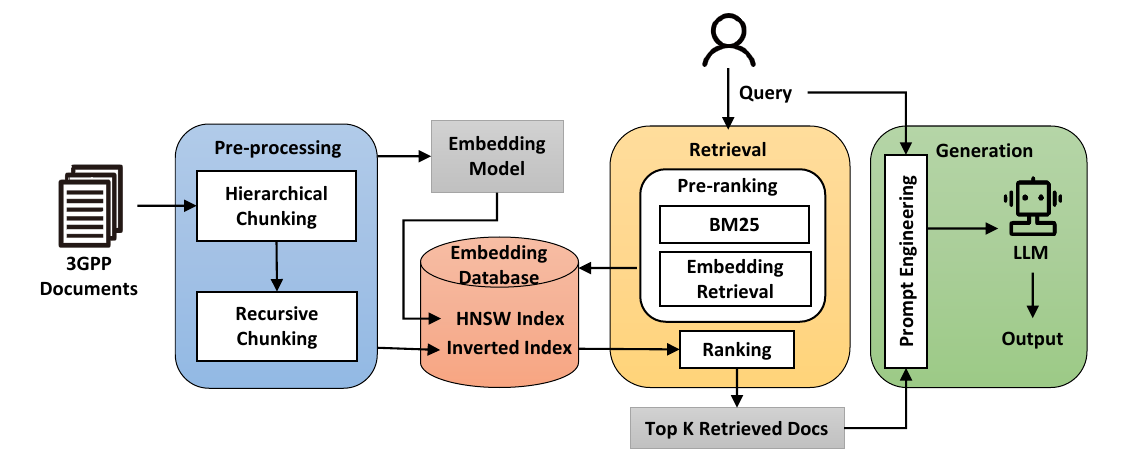}}
\caption{The overview of the proposed Chat3GPP.}
\label{fig}
\end{figure*}

\section{Methodology}
In this section, we introduce the methodology of Chat3GPP, as shown in Fig. \ref{fig}. The overall architecture involves several key phases, including data pre-processing, indexing, retrieval, and generation, ensuring both efficiency and accuracy in processing and retrieving information from the 3GPP technical documents.

\subsection{Data Pre-processing}
We first crawl the release 17 and release 18 technical specification documents from the 3GPP FTP site\cite{b15}. These documents provide detailed specifications for 3GPP technologies and serve as the primary knowledge source for Chat3GPP. We performed the following cleaning steps:
\par \textbf{Format Conversion.} We convert certain \textit{.doc} format documents into \textit{.docx} format in order to facilitate processing using the python-docx library.
\par \textbf{Content Filtering.}  We remov the \textit{Contents}, \textit{References}, and \textit{Annex} sections as these sections irrelevant to what we are concerned about.
\par \textbf{Text Extraction.} We extract the text content, excluding images and tables, ensuring that only textual data was used for further processing.

\subsection{Indexing Phase}
In the Indexing phase, documents will be processed, segmented, and transformed into vectorial representations to be stored in a embedding database. We use the following strategies including text segmentation, embeddings, and indexing strategy.
\subsubsection{Chunking Strategy}
Considering the context length limitations of the embedding model, we need to split them into smaller chunks. The quality of text chunks directly impacts the performance during the retrieval phase. Proper segmentation helps maintain both the document’s structural organization and its semantic integrity, ensuring that the retrieved results are both accurate and contextually relevant. To achieve this, we adopt the following chunking strategies:
\par \textbf{Hierarchical Chunking.} We split the specification document according to subheadings, ensuring that each text segment includes the full title path of its corresponding section to preserve the contextual information from the document's structure.
In the pre-ranking phase, we combine two retrieval methods: BM25-based retrieval \cite{b15} and embedding retrieval\cite{b16}.
\par \textbf{Recursive Chunking.} The most common chunking method is to split the document into chunks on a fixed number of tokens. However, token-based splitting method leads to truncation within sentences and depends on the tokenizer. Consequently, we used a splitter from langchain\cite{b17} named \textit{RecursiveCharacterTextSplitter} to split long texts into chunks of approximately 1250 characters without overlap, ensuring that the chunks fit within the context length limitation of the embedding model while preserving semantic integrity.

\subsubsection{Embedding}
In the RAG framework, embedding retrieval is performed by computing the similarity (e.g., cosine similarity) between the embeddings of the query and the text chunks. The effectiveness of this process heavily relies on the semantic representation power of the embedding model. For this purpose, we employed the open-source embedding model BGE-M3 \cite{18} to encode the document chunks into 1024-dimensional embedding vectors. These embeddings are then used for efficient retrieval in the subsequent stages.

\subsubsection{Indexing strategy}
The text chunks are stored in database in the structure of "filename, content, embedding", where:
\par \textbf{filename} represents the specification document of the text block, serving as a reference for tracing the source of the chunk.
\par \textbf{content} represents the natural language content of the text chunk, which is indexed as a \textit{text} type in Elasticsearch\cite{b19}. The \textit{text} field is stored using an inverted index, which allows efficient retrieval of documents containing specific query terms. To further improve retrieval efficiency, we customized the analyzer in Elasticsearch and employed a stop-word filter to remove irrelevant words, which helps reduce storage space and enhances search speed.
\par\textbf{embedding} represents the embedding vectors of the text blocks, stored in the \textit{dense\_vector} type and employing the Hierarchical Navigable Small World (HNSW) index. HNSW is an efficient approximate nearest neighbor (ANN) search algorithm that accelerates the search process through a graph-based approach.

\subsection{Retrieval Phase}
We employ a two-stage retrieval strategy, pre-ranking and ranking, to enhance both information retrieval efficiency and the quality of generated content as shown in Algorithm \ref{alg:al1}. In the pre-ranking phase, Chat3GPP first retrieves candidate chunks from the database that are relevant to the query using algorithms such as similarity-based retrieval. These candidate chunks are then scored for relevance, with the top-ranked chunks being selected for further processing, thereby filtering out irrelevant or low-quality candidates and improving retrieval efficiency. Considering both the input query and contextual requirements, the ranking stage further re-ranks candidate chunks to select the most helpful segments and provides high-quality input for the subsequent generation stage. Through the collaborative work of the pre-ranking and ranking stages, Chat3GPP can efficiently retrieve relevant information from large-scale databases and generate more accurate, relevant and high-quality answers.

\begin{algorithm}
\caption{Retrieval Strategies}
\label{alg:al1}
\begin{algorithmic}[1]  
\State Initialize the the embedding model for vectorial representations.
\State Initialize the 3GPP documents database.
\State Initialize the user query.
\State Pre-ranking:
\State \quad Filter chunks that contain query terms using the inverted \hspace*{1em}index, apply BM25 to score the chunks and retrieve the \hspace*{1em}Top-K1 chunks.
\State \quad Compute the cosine similarity between the query em-\hspace*{1em}bedding and chunk embeddings using HNSW and re-\hspace*{1em}trieve the Top-K1 chunks.
\State \quad Perform RRF to merge the rankings from BM25 and \hspace*{1em}embedding-based retrieval and return the \(\frac{1}{10}\) \(\text{Top-}K_1\) of \hspace*{1em}the combined chunks.
\State Ranking:
\State \quad For each chunk in the Top-K1 set, jointly encode the \hspace*{1em}query and chunk using the rerank model.
\State \quad Compute cosine similarity between the embeddings \hspace*{1em}of the query and each chunk, and return the Top-K2 \hspace*{1em}chunks as the final selection for subsequent generation.
\end{algorithmic}
\end{algorithm}

\subsubsection{Pre-ranking}
BM25\cite{b15} is a classic information retrieval algorithm that ranks documents by using the term frequency-inverse document frequency (TF-IDF) to calculate relevance between documents and the query. Specifically, BM25 calculates the relevance score of a document by considering the frequency of query terms within the document, while also accounting for document length and the frequency of the terms across the entire corpus. Although BM25 excels in keyword matching, it has limitations when it comes to understanding semantic meaning. For instance, BM25 fails to capture the relationships between words and struggles with complex query that require contextual understanding. As a result, embedding retrieval\cite{b16} has gradually emerged as a mainstream approach to assess the semantic relevance between documents and queries. In the Pre-ranking phase, we employ a hybrid retrieval approach that combines the strengths of both methods, ensuring that the most relevant documents are included in the candidate set.

First, we use Elasticsearch’s inverted index to filter out the text chunks that contain the query terms. Next, we apply the BM25 algorithm to score the remaining chunks, returning the \( \text{Top-}K_1 \) chunks in descending order of relevance. Afterward, we embed the user's original query using the same embedding model applied to the chunks without any query augmentation, and compute the cosine similarity between the query embedding and the chunks' embeddings using the HNSW index. The \( \text{Top-}K_1 \) documents are returned based on similarity scores.

To integrate the strengths of both retrieval methods, we employed Reciprocal Rank Fusion (RRF)\cite{b20} to reorder the chunks and return the top \(\frac{1}{10}\) of the Top-\(K_1\) chunks. RRF merges the ranking results from the two methods using a weighted combination, which not only deals with discrepancies in score computation, but also reduces biases that may exist in any single method and enhancing the accuracy and robustness of the combined results.

\subsubsection{Ranking}
In the ranking phase, we rerank of the chunks retrieved during the pre-ranking to prioritize the most relevant results using the rerank model. This refined selection provides more accurate inputs for subsequent processing by the language model, improving overall performance. Specifically, the BGE-M3 model, based on a cross-encoder architecture, jointly encodes the query and text chunks to generate new embedding vectors, computes cosine similarity between these embeddings and then the \( \text{Top-}K_2 \) documents are returned based on their similarity scores.

\begin{tcolorbox}[colframe=gray!80!black, colback=white, coltitle=black, title=\centering \textbf{Prompt for MCQs}]
You are an expert on 3GPP standards. Based on the provided context, answer the multiple-choice question by selecting the correct option.\\
Context:\{retrieved documents\}\\
Question:\{question\}\\
Options:\{options\}\\
Instructions:\\
1. Carefully review the context provided to determine the correct answer.\\
2. If the context does not provide sufficient information, respond with ``Insufficient context to answer.''\\
3. Provide your answer in the exact format: ``answer'': ``option X: [selected option content]''.\\
Answer:
\end{tcolorbox}

\begin{tcolorbox}[colframe=gray!80!black, colback=white, coltitle=black, title=\centering \textbf{Prompt for Open-ended questions}]
The following is a question about telecommunications and networking. Just give the answer based on the provided context.\\
Context:\{retrieved documents\}\\
Question:\{statement\}\\  
Answer:
\end{tcolorbox}

\subsection{Generation Phase}
During the generation phase, we employ a prompt engineering to incorporate the \( \text{Top-}K_2 \) retrieved text chunks along with the user’s query as input to the model. The model is then tasked with generating a response based on these documents. Prompt Engineering\cite{b21} refers to the process of designing and optimizing input prompts for LLMs to maximize the quality, relevance, and accuracy of the output. Therefore, we design distinct prompts specifically tailored for MCQs and open-ended questions to optimize performance across different query types.

\section{Experiments}
In this section, we use the settings for Chat3GPP as shown in Table \ref{tab:tab1}. Based on these settings,  We conduct experiments on two evaluation datasets to evaluate Chat3GPP framework in enhancing the functionality of LLMs applied to the telecommunications domain.

\begin{table}[htbp]
    \begin{center}
    \caption{SETTINGS OF CHAT3GPP}
    \resizebox{0.9\columnwidth}{!}{
    \label{tab:tab1}
    \begin{tabular}{cc}
    \toprule
    Parameter & Setting\\
    \midrule
    LLM & LLama3-8B-Instruct\\
    Embedding Model & BGE-M3\\
    Rerank Model & BGE-M3\\
    Embedding Database & Elasticsearch\\
    Indexing Strategy & Inverted Index/HNSW\\
    Chunk Size & about 1250 Characters\\
    Splitter & RecursiveCharacterTextSplitter\\
    \( \text{Top-}K_1 \) & 1000 \\
    \( \text{Top-}K_2 \) & 5 \\
    \bottomrule
    \end{tabular}
    }
    \end{center}
\end{table}

\subsection{Evaluation Datasets}
Currently, there are only two open-source evaluation datasets available to test the telecommunications knowledge of domain-adapted models: TeleQnA\cite{b4} and Tele-Eval\cite{b6}. TeleQnA is a MCQ dataset derived from standard and research papers. While MCQ datasets simplify accuracy evaluation, LLMs are particularly strong at MCQ selection due to the inherent ``selection bias'' common across nearly all LLMs\cite{b6}. To address limitation, \cite{b6} proposed an alternative approach by creating Tele-Eval, an open-ended telecommunications question dataset.

To assess the telecommunications capabilities of Chat3GPP, we used two evaluation sets. The first consists of 734 related to Release 17 and 780 related to Release 18 from TeleQnA. The second includes 26,225 questions related to Release 17 and 32,402 related to Release 18 from Tele-Eval.

\subsection{Evaluation Metrics and Benchmarks}
For the first evaluation dataset, users typically do not provide options when querying an LLM, so we excluded these options during the retrieval process. Instead, we incorporated into the prompt during the generation phase. In the following results as Table \ref{tab:tab2}, we use \textbf{accuracy} as the evaluation metric, measuring the proportion of correct answers provided by Chat3GPP for queries in the dataset. We selected TelecomGPT\cite{b5}, \textbf{Llama-3-8B-Tele-it}\cite{b6} which is the best-performing model from Tele-LLMs, and TelcoRAG\cite{b7} as benchmarks for comparison. Since TelecomGPT is a closed-source model, we directly adopted the benchmark results from the paper, which were evaluated on 3,500 questions, some of which derived from 3GPP documents, covering the \textit{Lexicon}, \textit{Standards Overview}, and \textit{Standards Specifications}.

\begin{table}[htbp]
    \begin{center}
    \caption{ACCURACY OF CHAT3GPP WITHOUT FINE-TUNING AND BENCHMARKS WITH FINE-TUNING}
    \resizebox{\columnwidth}{!}{
    \label{tab:tab2}
    \begin{tabular}{ccccc}
    \toprule
    Model & Rel.17 & Rel.18 & Overall &Finetuing \\
    \midrule
    TelecomGPT& - & - & 0.671 &Yes\\
    LLama3-8B-Tele-it & 0.531  & 0.571 & 0.552 &Yes\\
    Telco-RAG & 0.725 & 0.784 &- &Yes\\
    Chat3GPP & \textbf{0.783} & \textbf{0.791} & \textbf{0.787} & No\\
    \bottomrule
    \end{tabular}
    }
    \end{center}
\end{table}

The second evaluation dataset consists of open-ended question-answer pairs. Evaluating open-ended answers is more complex than MCQs. As noted in \cite{b5}, traditional evaluation metrics such as ROUGE\cite{b22} and BLEU\cite{b23} have limitations in this domain, and \textbf{LLM-Eval} is considered the most robust comparative tool for the telecommunications domain, which involves using an LLM as
a judge to assess the correctness of a model’s output compared to the ground truth answer. Thus, we used \textbf{LLM-Eval} as the evaluation metric and \textbf{Mixtral-8x7B-Instruct}\cite{b2} as the evaluation model to compare the model’s output with the ground truth and provide a Yes or No answer regarding its correctness using the same prompt as in \cite{b6}, and the results are shown in Table \ref{tab:tab3}. Additionally, we re-evaluated the second dataset using Llama-3-8B-Tele-it to ensure a more rigorous comparison of the results.

\begin{table}[htbp]
    \begin{center}
    \caption{LLM-EVAL RESULTS OF CHAT3GPP AND THE BENCHMARK}
    \resizebox{0.8\columnwidth}{!}{
    \label{tab:tab3}
    \begin{tabular}{cccc}
    \toprule
    Model & Rel.17 & Rel.18 & Overall\\
    \midrule
    LLama3-8B-Tele-it & 0.283 & 0.265 & 0.273\\
    Chat3GPP & \textbf{0.506} & \textbf{0.573} & \textbf{0.543}\\
    \bottomrule
    \end{tabular}
    }
    \end{center}
\end{table}

Overall, Chat3GPP demonstrated superior performance across all evaluation datasets. Despite being trained on telecom-specific data, specialized LLMs often perform inadequately when addressing issues related to 3GPP documents, showing the superiority of RAG framework in dealing with 3GPP documents. Furthermore, our RAG framework was not specifically trained on any model, while Telco-RAG trained an NN Router to predict relevant 3GPP documents, thereby reducing RAM usage.

It is worth mentioning that Telco-RAG\cite{b7} uses the FAISS index for retrieval, where all data must be loaded into RAM during the retrieval process. The memory consumption is directly proportional to the size of the dataset, resulting in significant memory usage, particularly with large-scale datasets. To mitigate this, Telco-RAG trained an NN Router that predicts the relevant 3GPP documents based on the query, selectively loading only the necessary embeddings, thus reducing RAM usage. In contrast, Elasticsearch minimizes memory consumption by employing an on-demand loading mechanism, ensuring that only the relevant data is loaded into RAM based on the query rather than the entire dataset.

\section{Conclusion}
In this paper, we introduce Chat3GPP, a RAG system specifically designed to enhance LLMs in understanding and applying 3GPP documents within the telecommunications domain. By leveraging the powerful combination of information retrieval and LLMs, Chat3GPP effectively addresses the challenges associated with the complexity, frequent updates, and specialized nature of 3GPP standards. This demonstrates the flexibility and scalability of the proposed framework, which can be easily adapted to other technical standards, such as those from ITU or IEEE. As a result, Chat3GPP presents a promising solution for improving the efficiency of engineers and technicians working with technical documentation. Additionally, it opens up new possibilities for tasks like document generation and code automation, further expanding its utility in the field.

However, several limitations remain. Firstly, future work could explore deeper integration between fine-tuned models and the retrieval-augmented framework to better understand their interactions and optimize the combination of retrieval and generation components. This could further enhance the model’s ability to respond to complex, domain-specific queries. Secondly, 3GPP documents contain not only textual data but also numerous tables and figures, which play a critical role in conveying information. Thus, integrating multi-modal data, such as tables and images, into the RAG framework could provide a richer, more contextually accurate foundation for text generation. In future work, we plan to explore further optimizations to better support a wider range of tasks within the telecommunications domain.

\section*{Acknowledgment}
This work was supported in part by National Natural Science Foundation of China (NSFC) under Grants 62394294, 62394290.


\end{document}